\documentclass{article}
\usepackage{spconf,amsmath,epsfig}

\usepackage{multirow}
\usepackage{array}
\usepackage{bbding}
\usepackage{pifont}
\usepackage{wasysym}
\usepackage{amssymb}
\usepackage{color}
\usepackage{booktabs}

\begin{document}\sloppy

\def\x{{\mathbf x}}
\def\L{{\cal L}}

\title{MFPN: A Novel Mixture Feature Pyramid Network of Multiple Architectures for Object Detection}
%
\name{Tingting Liang\textsuperscript{\rm 1}, Yongtao Wang\textsuperscript{\rm 1}, Qijie Zhao\textsuperscript{\rm 1}, huan zhang\textsuperscript{\rm 1}, Zhi Tang\textsuperscript{\rm 1}, Haibin Ling\textsuperscript{\rm 2}}
\address{\textsuperscript{\rm 1}Wangxuan Institute of Computer Technology, Peking University\\
 \textsuperscript{\rm 2}Department of Computer Science , Stony Brook University\\
 \{tingtingliang, wyt, zhaoqijie, zhanghuan666, tangzhi\}@pku.edu.cn\\
 hling@cs.stonybrook.edu\\ 
}

\maketitle

\begin{abstract}
Feature pyramids are widely exploited in many detectors to solve the scale variation problem for object detection. In this paper, we first investigate the Feature Pyramid Network (FPN) architectures and briefly categorize them into three typical fashions: top-down, bottom-up and fusing-splitting, which have their own merits for detecting small objects, large objects, and medium-sized objects, respectively. Further, we design three FPNs of different architectures and propose a novel Mixture Feature Pyramid Network (MFPN) which inherits the merits of all these three kinds of FPNs, by assembling the three kinds of FPNs in a parallel multi-branch architecture and mixing the features.  MFPN can significantly enhance both one-stage and two-stage FPN-based detectors with about 2 percent Average Precision(AP) increment on the MS-COCO benchmark, at little sacrifice in running time latency. By simply assembling MFPN with the one-stage and two-stage baseline detectors, we achieve competitive single-model detection results on the COCO detection benchmark without bells and whistles.
\end{abstract}

\section{Introduction}
\label{introduction}

Object detection is a fundamental research topic in image/video understanding. It can serve as a prerequisite for various image/video retrieval, intelligent surveillance and autonomous driving. Existing deep learning-based detectors can be briefly categorized into two branches: one-stage detectors such as SSD \cite{LiuAESRFB16}, RefineDet \cite{abs-1711-06897} and RetinaNet \cite{Lin2017}, which utilize CNN directly to predict the bounding boxes; and two-stage methods including Faster R-CNN \cite{RenHGS15}, R-FCN \cite{DaiLHS16} and Mask R-CNN \cite{He2017}, which generate a set of candidate proposals and then exploit the extracted region features from CNN for further refinement. Although encouraging progresses have been made, the existing detectors are still suffering from the problems caused by the scale variation across object instances.
\begin{figure}[t]
	\centering
	\includegraphics[scale=0.16]{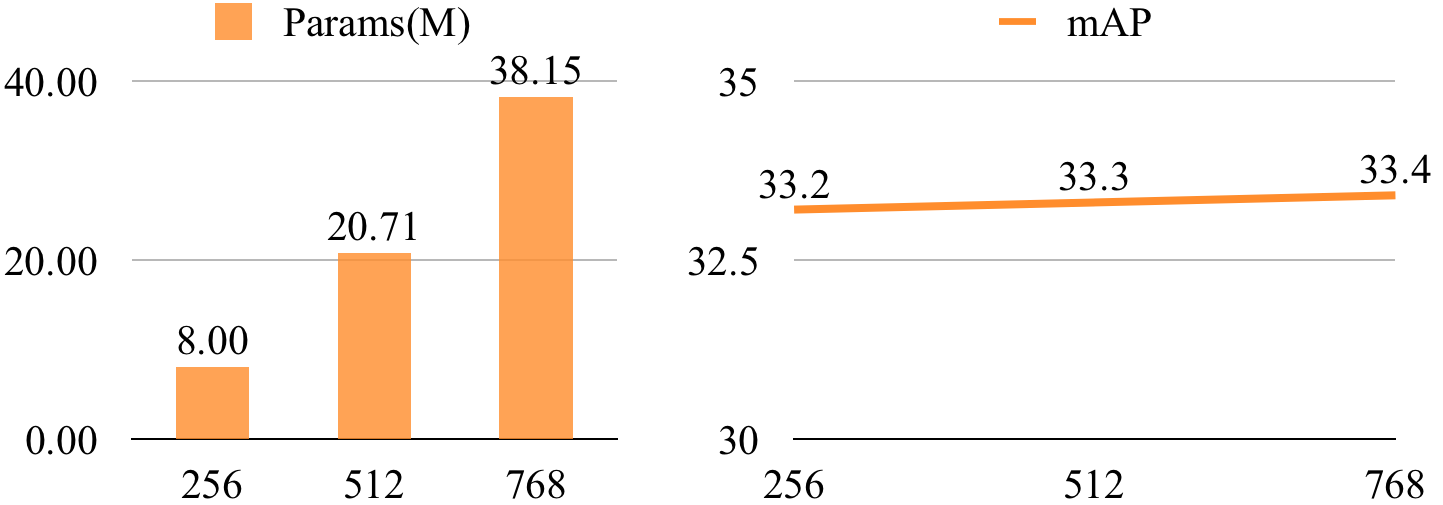}
	\caption{
		Left: the numbers of parameters with different numbers of channels of FPN. Right: the detection accuracies with different numbers of channels of FPN. The baseline detector is RetinaNet500-ResNet50.}
	\label{fpnparam}
\end{figure}
\begin{figure*}[t]
	\centering
	\includegraphics[scale=0.14]{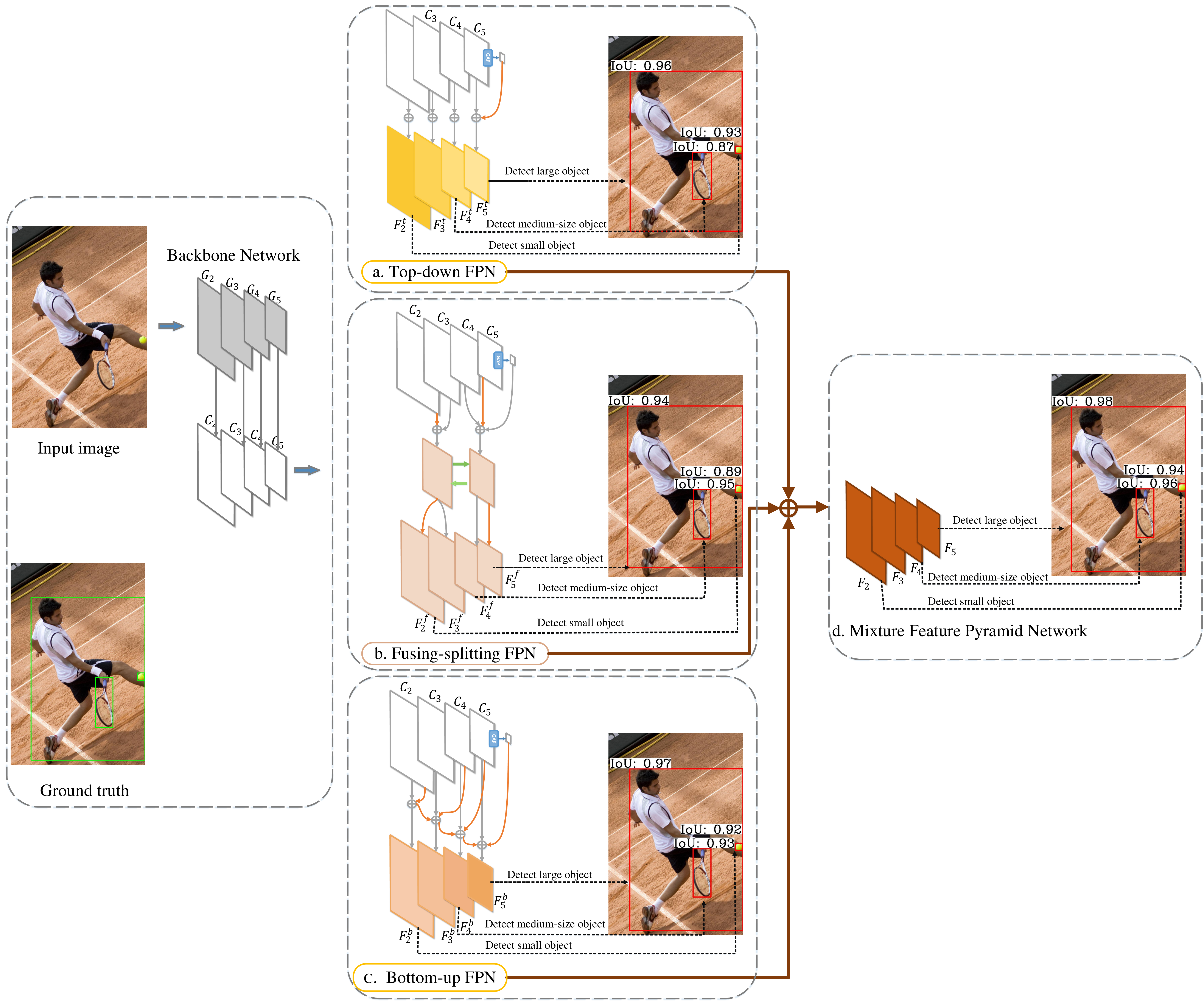}
	\caption{
		Exsample results of object detectors using feature pyramid networks of different architectures (the baseline detector is RetinaNet500-ResNet50). Our MFPN performs best: detecting objects of small-size, medium-size and large-size with the highest IoU. Green boxes: \textit{ground truth}, Red boxes: \textit{detection result}.}
	\label{shouye}
\end{figure*}
An intuitive approach to solve the scale variation problem is to use a multi-scale image pyramid \cite{abs-1711-08189}. However, the dramatic increase in inference time makes the image pyramid methods infeasible for practical applications. Other kinds of methods \cite{LinDGHHB17}\cite{ShrivastavaSMG16}\cite{KongSYLLC17}\cite{LiuAESRFB16} aim to employ the feature pyramid within the network, to approximate the image pyramid at a lower computational cost. Feature Pyramid Network (FPN) \cite{LinDGHHB17} is the most representative one, which incorporates high-semantic information in both high-level and low-level features with a top-down pathway, achieving superior performance. However, this top-down architecture design has the following intrinsic limitations: (1) it only introduces high-semantics information from deep layer to shallow layer, but does not consider the assistance of shallow layer to deep layer; (2) the top-down architecture makes the features of small objects largely depend on the features of larger objects, and this dependence is not always beneficial. 
For instance, we conduct a toy experiment by change the number of FPN channels in the baseline detector RetinaNet-ResNet50 (input size 800) \cite{Lin2017} to test the accuracy bottleneck, and the results are shown in Figure \ref{fpnparam}. It is notable that when the channel dimension increases to 768,  the accuracy growth is negligible with a lot of additional computation and parameters. This experiment demonstrates that such a top-down FPN architecture has bottleneck restrictions.

To address these problems, we rethink the feature pyramid network and summarize the architectures of FPN with three different fashions: top-down, fusing-splitting and bottom-up. As illustrated in Figure \ref{shouye} from top to bottom, we design an instance FPN for each FPN architecture. The Top-down FPN is an improved version of the original FPN \cite{LinDGHHB17}, which introduces high-level semantic contexts to low-level features for better detecting small objects. In particular, we newly propose the bottom-up FPN, which introduces low-level details to high-level features, helping the high-level features obtain more spatial information thus can better detect large objects. Deviated from the interdependent relationship between deep and shallow features, we propose a novel Fusing-splitting FPN, which first fuses higher-level and lower-level features and then splits the fused feature into multi-scale features. Further, as illustrated in Figure \ref{shouye}, we propose a novel feature pyramid network that assembles these three FPNs of different architectures, named Mixture Feature Pyramid Network (MFPN).
Experimental results show that the proposed MFPN can significantly enhance these FPN based detectors by about 2 percent Average Precision(AP), and can improve the detection performance of objects of all scale ranges (e.g., as depicted in Figure \ref{shouye}). Moreover, competitive single-model detection results are achieved by both one-stage and two-stage baseline detectors equipped with MFPN.

In summary, our main contributions are as follows:
\begin{itemize}
	\setlength{\itemsep}{1pt}
	\setlength{\parsep}{0pt}
	\setlength{\parskip}{0pt}
\item We design three FPNs of different architectures, Top-down FPN, Bottom-up FPN, and Fusing-splitting FPN, which have better detection performance for small objects, large objects, medium-size objects respectively.
\item We propose a novel Mixture Feature Pyramid Network (MFPN) which inherits all the merits of the three FPNs, by assembling them in a parallel multi-branch architecture and mixing the features extracted by each branch.
\item We achieve significant better detection results than both one-stage and two-stage FPN-based detectors on MS COCO benchmark.
\end{itemize}

\section{Related Work}

Addressing scale variation issue is critical for object detection, segmentation and other tasks that require predictive location\cite{abs-1711-08189}. 
To tackle the scale variation problem, an intuitive way is to use a multi-scale image pyramid during training and inference \cite{DaiLHS16}\cite{DaiQXLZHW17}\cite{HuangRSZKFFWSG017}.  Different from  methods with fixed or random scale transform, SNIP \cite{abs-1711-08189} selectively back-propagates the gradients of object instances of different sizes as a function of image scale. In addition, SNIPER \cite{Singh2018} samples low-resolution chips to accelerate multi-scale training. Multi-scale image pyramid greatly improves accuracy, but suffers a lot from increasing inference time. 

The feature pyramid method, that is, constructing and using the feature pyramid within the network, is more widely used to deal with scale variation, due to its lower computation cost. Methods like SSD \cite{LiuAESRFB16} and MS-CNN \cite{Cai2016} directly perform small objects detection on higher resolution feature maps while large ones on lower resolution feature maps extracted by the backbone network (e.g., VGG). Due to the backbone networks are originally designed for classification task, directly using the features extracted by them leads to suboptimal performance. Hence, some recent works try to alleviate this problem by enhancing the features extracted by backbones with novel feature enhancement modules, e.g., RFBNet \cite{LiuHW18} and TridentNet \cite{Li2019}.
Feature Pyramid Networks (FPN) \cite{LinDGHHB17} is commonly exploited by state-of-the-art object detectors, e.g., Mask RCNN \cite{He2017}, RetinaNet\cite{Lin2017}, RefineDet \cite{abs-1711-06897}, etc., which proposes a subnet with top-down architecture to construct feature pyramid. Recently, Multi-level FPN\cite{zhao2019m2det} introduces multiple U-shape modules after a backbone network to extract multi-level pyramidal features, and builds a powerful one-stage detector. Libra R-CNN \cite{Pang2019} and \cite{Kong2018} are two recently proposed feature pyramid networks of Fusing-splitting architecture, who combine features of all scales and then generate features at each scale by a global attention operation on the combined features. As stated in section \ref{introduction}, these FPNs have their own intrinsic limitations since they are designed with only one specific kind of FPN architecture (i.e., top-down, or fusing-splitting, or bottom-up).

\section{Proposed Method}
\label{sec:3topologies}
In this work, we first introduce three kinds of FPN architectures, that is, Top-down, Bottom-up and Fusing-splitting. As illustrated in Figure \ref{shouye}, each pyramidal feature map (denoted as $G_2, G_3, G_4, G_5$) extracted by the backbone is followed by an extra $1\times1$ convolution. Then, these feature maps (denoted $C_2, C_3, C_4, C_5$) are used to build feature pyramid for object detection by each FPN of different architectures as following.

\subsection{Top-down FPN}
The major characteristic of top-down FPN architecture is: the FPN feature maps (denoted as $F_2^{t},F_3^{t},F_4^{t},F_5^{t}$) are sequentially constructed in a top-down manner, that is, the smaller scale (higher-level) feature map is constructed first.
we adopt the most widely used top-down architecture FPN\cite{LinDGHHB17} with some modifications. To be more specific, we plug an extra global average pooling(GAP)\cite{LinCY13} layer above the deepest layer of the backbone to extract the global context, i.e., G5. Moreover, GAP can learn richer semantic information and highlights the discriminative object regions detected by CNNs\cite{SzegedyLJSRAEVR15}, thus can propagate more semantic information to the larger scale(lower-level) feature maps. Same as the original FPN\cite{LinDGHHB17}, each feature map ($F_i^{t}$) of Top-down FPN is iteratively built by combining the same level backbone feature map ($C_i$) and the higher-level FPN feature map ($F_{i+1}^{t}$):

\begin{equation}
{
	F_i^{t}=\mathbf{W^t}_{i+1}\otimes(\textit{U}(F^t_{i+1})+C_{i}),
}
\label{eq:fpn}
\end{equation}
where $\textit{U}(\cdot)$ denotes the upsample operation with a factor of 2 and $\mathbf{W_i^t}$ is a $3\times3$ convolution filter. Since the top-down architecture iteratively propagates semantic information of higher-level backbone features to the more detailed lower-level FPN feature maps, it is better at detecting small objects.

\subsection{Bottom-up FPN}
Contrary to the top-down architecture, the major characteristic of bottom-up FPN is: the FPN feature maps (denoted as $F_2^{b},F_3^{b},F_4^{b},F_5^{b}$) are sequentially constructed in a bottom-up manner, that is, the large scale (lower-level) feature map is constructed first. As illustrated in Figure \ref{shouye}.c, each feature map ($F_i^{b}$) of the Bottom-up FPN is obtained by merging the same level backbone feature map ($C_{i}$), the backbone feature map ($C_{i+1}$) above it, and the FPN feature map ($F_{i-1}^{b}$) below it, which can be formulated as:

\begin{equation}
{
	\begin{aligned}
	F_{i}^{b}=\mathbf{W_{i}^b}\otimes(\textit{D}(F^b_{i-1})+C_i+\textit{U}(C_{i+1})),
	\end{aligned}
}
\end{equation}
where $\textit{D}(\cdot)$ denotes MaxPool operation with a factor of 2 and $\mathbf{W_i^b}$ is a $3\times3$ convolution filter.  Because the bottom-up architecture propagates the spatial detail information of lower-level backbone features to the higher-level FPN features, it is better at detecting large objects. Obviously, Bottom-up FPN and Top-down FPN are complementary to each other. 

\subsection{Fusing-splitting FPN}

Since the feature maps of the Top-down FPN and Bottom-up FPN are sequentially built, the earlier constructed features always affect the subsequent ones, and this interdependent design may lead to some intrinsic limitation. To address this problem, we design a Fusing-splitting FPN, which first combines the higher-level and lower-level backbone features, and then splitting the combined features to multi-scale FPN features. 
\begin{table}[hbt]
	\centering
	\small

	\caption{
		Object detection result comparison on COCO minival for the three FPNs of different architectures and the proposed MFPN. The baseline is RetinaNet500-ResNet50.}
	\vspace{0.2cm}
	\begin{tabular}
		{lcccccc}
	
		\toprule
		\textbf{Method} & \textbf{Parameters(M)}  &\textbf{AP}  & \textbf{$AP_{s}$} & \textbf{$AP_{m}$} & \textbf{$AP_{l}$}  \\ 
		\midrule
		FPN (Baseline) & 8.00   & 33.2 & 15.0  & 37.5 & 47.4\\
		\hline
		Top-down  & 8.52  &33.5 & \textbf{15.2} & 38.1  & 47.6\\
		Bottom-up  & 8.52  &33.5 & 14.4 & 37.9 & \textbf{48.7}\\
		Fusing-splitting & 6.49  &33.6 & 14.7 & \textbf{38.5}  & 48.1\\
		\hline

		MFPN  & 11.47 & \textbf{34.8} & \textbf{16.8} & \textbf{39.1}  & \textbf{49.0}\\
		
		\bottomrule 
	\end{tabular}
	
	\label{detection result 1} 
\end{table}
\begin{table}[!tbh]
	\centering
	\small
	\caption{Object detection results comparison on COCO minival for different combinations of the three kinds of FPN architectures. The baseline is RetinaNet500-ResNet50.}
	\vspace{0.2cm}
	\begin{tabular}
		{p{3.8cm}p{0.5cm}<{\centering}p{0.5cm}<{\centering}p{0.5cm}<{\centering}p{0.5cm}<{\centering}p{0.5cm}<{\centering}p{0.5cm}<{\centering}}
		\toprule
		\textbf{Method} &  \textbf{AP}  & $AP_{s}$ & $AP_{m}$ & $AP_{l}$\\ 
		\midrule
		Baseline   &33.2 
		&15.0  & 37.5 & 47.4\\
		\hline
		Bottom-up + Fusing-splitting  &34.3 
				 &16.0 &39.0 &48.6\\
		Top-down + Bottom-up  &34.3
			&15.8 &38.9 &48.9\\
		Top-down + Fusing-splitting  &33.8
			&15.7 &38.4 &47.7\\
		MFPN 	   &\textbf{34.8} & \textbf{16.8} & \textbf{39.1}  & \textbf{49.0}\\
		\bottomrule
	\end{tabular}
	
	\label{detection result 7} 
\end{table}
In practice, the highest two backbone feature maps are merged into a combined feature map $\alpha_s$, and the lowest two backbone feature maps are merged into  $\alpha_l$:

\begin{equation}
{
	\alpha_s=C_4+\textit{U}(C_{5}), 
	\alpha_l=\textit{D}(C_2)+C_{3}.
}
\end{equation}
After obtaining the first-round combined features, we further fuse them as following, 
\begin{equation}
{
	\begin{aligned}
	\beta_s=\mathbf{W_{s}^f}\otimes cat(\alpha_s, \textit{D}(\alpha_l)),
	\\
	\beta_l=\mathbf{W_{l}^f}\otimes cat(\textit{U}(\alpha_s), \alpha_l),
	\end{aligned}
}
\end{equation}
where $\mathbf{W_{s}^f}$ and $\mathbf{W_{l}^f}$ are two $3\times3$ convolution filters, and $cat(\cdot)$ represents concat operation along channel dimension. 
After these operations, feature maps $\beta_s, \beta_l$ have fused informations from all level features. Finally, we simply resize $\beta_s, \beta_l$ into multi-scale pyramidal feature maps, that is, 

\begin{equation}
{
	\begin{aligned}
	&F_2^{f}=\textit{U}(\beta_l),
	F_3^{f}=\beta_l;\\
	&F_4^{f}=\beta_s,
	F_5^{f}=\textit{D}(\beta_s).
	\end{aligned}
}
\end{equation}
By the above two rounds fusing and the splitting operations, all the feature maps of the Fusing-splitting FPN incorporate information from the backbone feature maps of all levels. Moreover, \textit{the two medium-scale feature maps} ($F_3^{f}$ and $F_4^{f}$) are obtained with less downsampling or upsampling operation. Hence, Fusing-splitting FPN has a stable improvement in detecting medium-sized objects.
\subsection{Mixture Feature Pyramid Network (MFPN)}

\begin{table}[!tbh]
	\centering
	\small
	\caption{Performance comparison between FPN and MFPN on the  COCO minival. R: ResNet. X: ResNext-101-64x4d.}
	\vspace{0.2cm}
	\begin{tabular}
		{p{2.4cm}p{0.8cm}<{\centering}p{0.33cm}<{\centering}p{0.3cm}<{\centering}p{0.3cm}<{\centering}p{0.3cm}<{\centering}p{0.975cm}<{\centering}}
		\toprule
		\textbf{Baseline} &\textbf{Method}  &$AP$    & $AP_{S}$ & $AP_{M}$ & $AP_{L}$ & \textbf{time(ms)} \\
		\midrule
		Retinanet-R50   \multirow{2}{*}&FPN &35.6	&20.0	&39.6	&46.8&85\\
		&ours	&37.9 &21.4 &41.9 &49.7&86\\\hline
		Retinanet-X101   \multirow{2}{*}&FPN &40.0	&23.0	&44.3  &52.7&196\\
		&ours	&42.1  &24.9 &46.8 &55.3&196\\\hline
		Faster R-CNN  \multirow{2}{*}&FPN &36.4&21.5&40.0&46.6&82\\
		-R50&ours	&38.6&22.6&42.8&49.7&93\\\hline
		Cascade Mask R-\multirow{2}{*}&FPN &42.7	&23.8  &46.5&56.9&196\\
		CNN-R101 &ours	&44.4	&25.9	&48.1	&58.2&204\\
		\bottomrule
	
	\end{tabular}
	
	\label{detection result 10} 
\end{table}

Now we propose a more powerful feature pyramid network named MFPN by integrating the above three FPNs. Intuitively, MFPN inherits all the merits of the three FPNs and performs better to handle scale variation problem in object detection. By integrating the three FPNs in one network, we can avoid a large increase in the number of parameters by sharing one backbone network. 
The network architecture of MFPN is illustrated in Figure \ref{shouye},
each feature map of MFPN is obtained by summing the same level feature map of the three feature pyramids along spatial dimension, that is,
\begin{equation}
F_i = F_i^t + F_i^b + F_i^f, i=2,3,4,5.
\end{equation}
MFPN can play all the roles played by FPN, including as anchor feature to improve the accuracy\cite{Lin2017}, or as neck feature to boost RPN\cite{RenHGS15} for better candidate proposals and connect with RoI Extractor\cite{RenHGS15}\cite{DaiLHS16}\cite{He2017}for better RoI features.

\section{Experiment}
\begin{table*}[!tbh]
	\centering
	\small
	\label{tab:cocoresults}
	\centering
	\caption{Detection accuracy comparisons with the state-of-the-art FPN-based methods on \textit{MS-COCO} \texttt{test-dev} set.}
	\begin{tabular}{l|c|cccccc}
		\toprule	
		\textbf{Method} & \textbf{Backbone} & \textbf{AP} & AP$_{50}$ & AP$_{75}$ & AP$_{s}$ & AP$_{m}$ & AP$_{l}$ \\
		\midrule
		\small\textit{\textbf{one-stage:}} & & & & & & &\quad \\
		SSD512 \cite{LiuAESRFB16}  & VGG-16 & 28.8 & 48.5 & 30.3 & 10.9 & 31.8 & 43.5 \\
	
		RefineDet512 \cite{abs-1711-06897} & ResNet-101 & 36.4 & 57.5 & 39.5 & 16.6 & 39.9 & 51.4 \\
		RetinaNet800 \cite{Lin2017}  & Res101-FPN & 39.1 & 59.1 & 42.3 & 21.8 & 42.7 & 50.2 \\
		CornerNet \cite{abs-1808-01244} & Hourglass-104 & 40.5&56.5&43.1&19.4&42.7&53.9\\
		M2Det \cite{zhao2019m2det} &VGG-16  & 41.0 & 59.7 & 45.0 & 22.1 & 46.5 & 53.8 \\
		FSAF \cite{DBLP:conf/cvpr/ZhuHS19} &ResNext-101-64x4d &42.9 &63.8 &46.3 &26.6 &46.2 &52.7\\
		
		\hline
		
		\small\textit{\textbf{two-stage:}} & & & & & & &\quad \\

		Faster R-CNN w FPN \cite{LinDGHHB17}  & ResNet101-FPN & 36.2 & 59.1 & 39.0 & 18.2 & 39.0 & 48.2 \\
		Deformable R-FCN \cite{DaiQXLZHW17}  & Inc-Res-v2 & 37.5 & 58.0 & 40.8 & 19.4 & 40.1 & 52.5 \\
		Mask R-CNN \cite{He2017} &ResNeXt-101 & 39.8 & 62.3 & 43.4 & 22.1 & 43.2& 51.2\\
		TridentNet \cite{Li2019} &ResNet-101-Deformable  & 42.7 & 63.6& 46.5 & 23.9 & 46.6& 56.6\\
		Cascade R-CNN \cite{abs-1712-00726} &ResNet101-FPN & 42.8 &62.1&46.3&23.7&45.5&55.2\\
		SNIP \cite{abs-1711-08189} &ResNet-101-Deformable  & 44.4 & 66.2 & 44.9 & 27.3 & 47.4 & 56.9 \\
		SNIPER \cite{Singh2018} &ResNet-101-Deformable   & 46.1 & 67.0 & 51.6 & 29.6 & 48.9 & 58.1\\

		\hline
		\small\textit{\textbf{Ours:}} & & & & & & &\quad \\
		
		MFPN-Cascade Mask R-CNN&ResNext-101-64x4d&47.6 &66.7 &52.0 &29.4 &50.8 &59.6\\
		MFPN-RetinaNet & ResNext-101-64x4d &43.4 &63.4 &46.5 &26.1 &47.3 &54.0\\

		\bottomrule
	\end{tabular}
	
	\label{detection result 8}
\end{table*}

\subsection{Dataset and Implementation details}

\textbf{Dataset Description.} 
We present experimental results on the bounding boxes detection task of the challenging \textit{MS-COCO} benchmark \cite{LinMBHPRDZ14}. For training, validation and testing processes, we follow \cite{abs-1711-06897} and \cite{LinDGHHB17}, train on the union of 11.8k training images(including the 80k \texttt{train} split and a random 35k subset of images from the 40k image \texttt{val} split), conduct ablation study on 5k \texttt{minival} split for convenience. Then, to compare the accuracy with state-of-the-art FPN-based methods, we report results of \texttt{test-dev} split images.

\vspace{1mm}
\noindent\textbf{Implementation Details.} The backbones used in this paper are all pre-trained on ImageNet \cite{RussakovskyDSKS15}.
For ablation study experiments, we train detectors 12 epochs in total, with learning rate starting from 0.02 and the batch size is 16.  Cascade Mask R-CNN-MFPN and RetinaNet-X101-MFPN are trained for 20 epochs and the initial learning rate is set to 0.01.  For evaluation, detectors run on a single Titan X GPU with CUDA 9 and CUDNN 7, with a batch size of 1.

\subsection{Ablation Studies}

\textbf{Compare the three FPNs} 
\begin{figure}[bht]
	\centering
	\includegraphics[width=\linewidth]{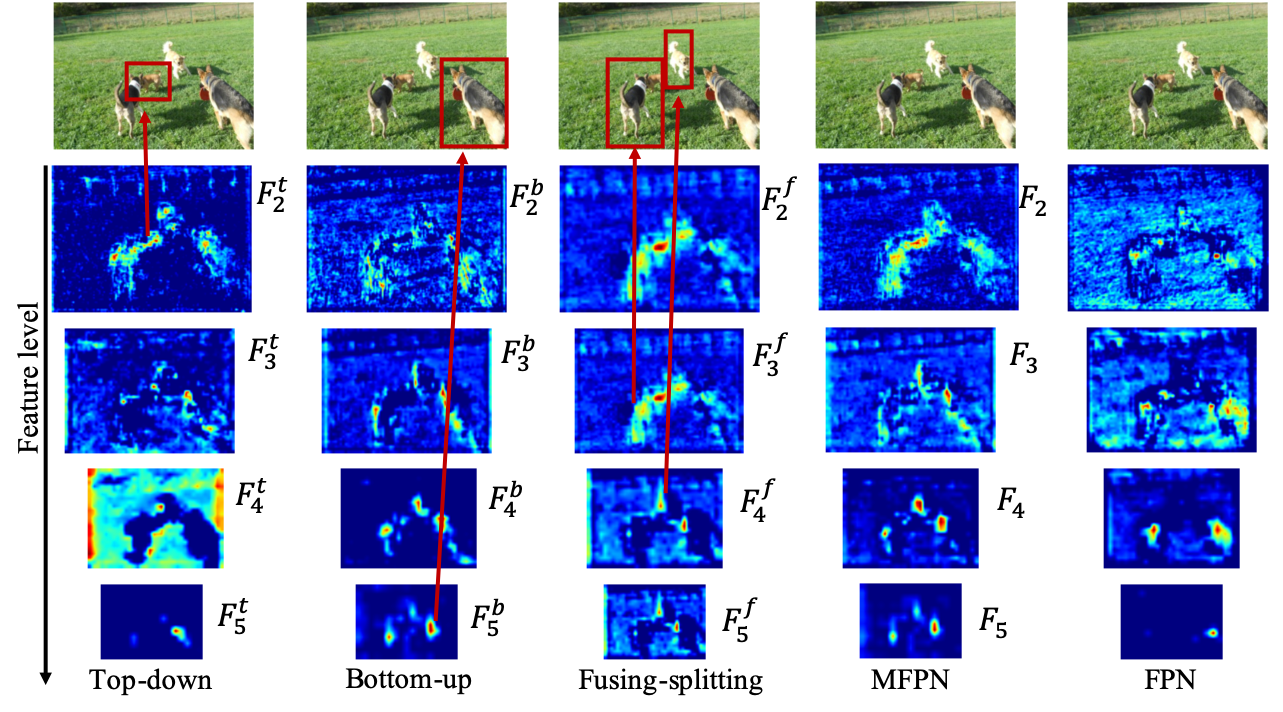}
	\caption{Heatmap visualization exsamples of MFPN and FPN.}
	\label{visualize}
\end{figure}
As shown in Table~\ref{detection result 1}, Top-down FPN gets the highest score for small objects($AP_s$ of 15.2), while Bottom-up FPN wins for large objects($AP_l$ of 48.7) and Fusing-splitting FPN is best at detecting medium-sized objects($AP_m$ of 38.5). When we add up the three FPNs, the overall AP is 1.5 higher than FPN. 
We also conduct experiments of multiple combinations of Top-down FPN, Bottom-up FPN and Fusing-splitting FPN in Table~\ref{detection result 7}. The combination of Top-down and Bottom-up gets the highest result (36.8) among the pair-wise combinations. At the same time, to further improve the accuracy of AP$_{75}$ and enhance the detection accuracy of hard samples, we adopt a combination of three FPNs. 
These results fully confirm our expectations and prove that our design is reasonable and effective.

\vspace{1mm}
\noindent\textbf{MFPN can significantly enhance FPN-based detectors} We further evaluate the proposed MFPN with different backbones and detectors, using input image scale of 800 pixels. Results are detailed in Table \ref{detection result 10}. MFPN consistently improves the detection accuracy for various backbones. 
For MFPN-Retinanet and MFPN-Faster R-CNN, we adopt balanced loss\cite{Pang2019} instead of smooth L1 to better handle sample imbalance problem.
Our MFPN introduces marginal computation cost to the whole detection network, leading to negligible loss of inference speed. Especially, we improve RetinaNet by 2.1 AP on Retinanet ResNeXt-101 without additional inference latency increment, and 1.6 percent of AP on Cascade Mask RCNN-ResNet 101 with only 8ms latency increment. 

\vspace{1mm}
\noindent\textbf{MFPN can learn better features for object detection}
To verify that the proposed MFPN can learn effective feature for detecting objects of various sizes, we visualize the activation values of the output of FPN and MFPN along scale and level dimensions, such an example shown in 
Figure~\ref{visualize}.  The input image contains four dogs with different sizes. We can find that: 1) For detecting the smallest dog, the lowest feature from Top-down FPN $F_2^t$ achieves clearer and noise-free semantics than that from FPN. 
2) Compared with FPN, Bottom-up FPN obtains better high-level FPN features with three clear activation points in $F_5^b$, and can better detecting the biggest dog.
3) $F_3^f,F_4^f$ from Fusing-splitting FPN have larger activation regions than FPN, containing more detailed information, thus cann better detecting the two medium-sized dogs. 4) The responses of MFPN to objects are accurate, while the ones of FPN are hindered by meaningless noise. This implies: 1) MFPN is good at learning the characteristics of objects. 2) It is necessary to use MFPN to detect objects of various sizes.



\subsection{Compare with state-of-the-art FPN-based methods}
We evaluate MFPN on the \textit{COCO} \texttt{test-dev} set and compare it with recent state-of-the-art FPN-based methods.
The model is trained using scale jitter over scales $\{640, 672, 704, 736, 768, 800\}$.
For fair comparison, we only compare the results produced from single models without ensemble or multi-scale testing.
As shown in Table~\ref{detection result 8},
MFPN based detectors, RetinaNet-MFPN, and Cascade Mask R-CNN-MFPN, achieve superior results without bells and whistles. RetinaNet-MFPN gets AP (\textbf{43.4}), which surpasses all other one-stage detectors. Cascade Mask R-CNN-MFPN obtains AP of \textbf{47.6}, outperforms TridentNet, SNIP and SNIPER, who uses image pyramid training and testing strategies. In conclude, MFPN is compatible with both powerful one-stage detectors and two-stage detectors and can achieve very competitive single-model results.

\section{Conclusion}

In this paper, we first describe three FPNs of different architectures(\textit{i.e.,}, Top-down, Bottom-up, and Fusing-splitting) for extracting multi-scale features to solve the scale variation problem for object detection. Based on them, we propose a novel Mixture Feature Pyramid Network(MFPN), which is effective for learning powerful multi-scale features and can be simply assembled into both one-stage detectors and two-stage detectors. On the MS-COCO benchmark, MFPN improves the performance for all scale-ranges and enhances both one-stage and two-stage FPN-based detectors with 2 \% AP increment, which leads to very competitive results.

\bibliographystyle{IEEEbib}
\bibliography{detection}

\end{document}